\documentclass[sigconf]{acmart}

\AtBeginDocument{%
  }

\setcopyright{acmlicensed}
\copyrightyear{2025}
\acmYear{2025}
\acmDOI{10.1145/3721146.3721958}
\acmISBN{979-8-4007-1538-9/25/03}

\acmYear{2025}\copyrightyear{2025}
\acmConference[EuroMLSys '25]{The 5th Workshop on Machine Learning and Systems}{March 30--April 3, 2025}{Rotterdam, Netherlands}
\acmBooktitle{The 5th Workshop on Machine Learning and Systems (EuroMLSys '25), March 30--April 3, 2025, Rotterdam, Netherlands}

\PassOptionsToPackage{dvipsnames}{xcolor}
\begin{document}

\title{Diagnosing and Resolving Cloud Platform Instability
with Multi-modal RAG LLMs}

\author{Yifan Wang}
\email{wangyifan@cs.cornell.edu}
\orcid{0009-0004-6747-2229}
\affiliation{%
  \institution{Computer Science Department, Cornell University}
  \city{Ithaca}
  \state{NY}
  \country{USA}
}

\author{Kenneth P. Birman}
\email{ken@cs.cornell.edu}
\orcid{0000-0003-2400-149X}
\affiliation{%
  \institution{Computer Science Department, Cornell University}
  \city{Ithaca}
  \state{NY}
  \country{USA}
}

\renewcommand{\shortauthors}{Wang et al.}

\begin{abstract}
Today's cloud-hosted applications and services are complex systems, and a performance or functional instability can have dozens or hundreds of potential root causes.  Our hypothesis is that by combining the pattern matching capabilities of modern AI tools with a natural multi-modal RAG LLM interface, problem identification and resolution can be simplified. ARCA is a new multi-modal RAG LLM system that targets this domain. Step-wise evaluations show that ARCA outperforms state-of-the-art alternatives.
\end{abstract}




\received{11 February 2025}
\received[accepted]{25 February 2025}
\received[revised]{7 March 2025}

\maketitle

\section{Introduction}
Incident response in complex systems entails 4 steps. (1) {\em Detection,} which includes the detection or prediction of an impending problem; (2) {\em Triage:} categorizing severity and assigning the task to a Site Reliability Engineering (SRE) team; (3) {\em Diagnosis:} collecting more data and pinpointing the root cause; (4) {\em Mitigation:} Formulating and carrying out a response and disabling any extra instrumentation that was activated.

Decades of work has given us a remarkable range of AI-assisted IT-Operation (AI-Ops) tools covering each step, such as prediction-based anomaly alarming, classification-based internal support ticket assigning tool for triaging, root-cause analysis tools using language models for summarization and many more. These AI tools work on a variety of data modalities, including user-provided bug reports in natural language, system logs in a semi-structured language and numerical performance metrics.

% multi-modality is hard <-> ARCA: multi-modal solution
Our work takes the next step by offering an AI-Ops solution that can carry out cross-modality reasoning. 
%Even when data sets are purely numerical, proper interpretation of the cloud instability will depend on the system component being monitored, the semantics of the monitoring output, correlations with logs, traces, software and hardware update records, etc. 
The task is challenging for several reasons: multi-modal language models are still in an a very early stage, and there is a lack of a significant lack of high-quality training data sets for our setting. To the extent that one can identify public data sets for AI-Ops and IT-Ops they generally offer just a single data mode, as is the case for the two most widely cited sets, HPC4 \cite{HPC4}, COM2 \cite{com2}.  But this issue is also seen with less widely used data sets.

% non-resilliency <-> ARCA: unsupervised approach, similarity search
Even if we limit ourselves to a single data mode, existing AI-Ops solutions turn out to have limitations (such as weak support for events characterized by evolution of a problem over time, and hence recognizable only from a series of log records), and also struggle to adapt to changes in their operating environment. If the underlying data distribution shifts, for example after a hardware upgrade, the performance of threshold-based incident detection tool is often found to degrade. Upgrades often result in logging new information, yet small modifications in the log formatting can defeat log analytics implemented with regular expressions.  As a result, users of today's solutions complain about frequent forced code changes and the need for periodic model retraining. %Such rigidity is introduced at the birth of the tool when the training data can only cover limited scenarios and it cannot be future-proof. Hence lots of SRE efforts are spent on aligning the tools with new settings.

% no general tool and expensive <-> ARCA: off-the-shelf model, financially reasonable solution
Beyond these technical limitations, today's AI-Ops tools are often proprietary and forbiddingly expensive.  DevOps teams at cloud computing companies with vast GPU deployments can train new models, but this is out of the question for smaller companies.  % Many new tools gain their efficacy by training or fine-tuning proprietary Large Language Models (LLM) for their internal tasks. However, such practice requires not only heavy investment on accelerators like GPUs, but also an abundant amount of private training data that takes years of data collection. Neither of those are affordable by new practitioners and they become the hidden barrier only benefiting the big businesses.

% Intuition: multi-modal solution
%We see the need for resiliency, generic-ness, and multi-modal reasoning capability as a match with {\em Retrieval-Augmented Generative} (RAG) LLMs, in which an LLM is augmented with a database that could support many data modalities. 
ARCA is an {\em AI for Root Cause Analysis} based on a multimodal RAG (Retrieval-Augmented Generation) approach, in which an LLM is augmented by a database.  Many RAG systems are limited to approximate search in document or image collections, but ARCA also supports data in structured (tabular) collections and logs. The basic idea is to focus on recurrent incidents, looking for similar past problems, summarizing prior findings, and recommending mitigation strategies that succeeded in the past.   

% resillient 
A complicating factor is that users often report incidents in fuzzy ways, which limits label quality: a particular problem given that many AI-Ops tools are trained using labeled data. Rather than battling this reality, our work focuses on {\em approximate match} (an idea familiar in text-based contexts), but generalizes the mechanism to to encompass data modalities other than text. The idea, though, is similar: RAG LLMs for search document collections treat each query as a vector database search for documents ``similar'' to the query. ARCA treats the multimodal signature of the incident as a kind of query and performs  approximate match  against precomputed signatures from past incidents.

% multi-modal
Here we report on a proof-of-concept that supports three data modes: (1) {\em incident descriptions},  in natural language; (2) {\em logs} of semi-structured text generated by automated reporting components; and (3) multivariate performance-counter time-series.  ARCA is an end-to-end tool created from off-the-shelf ML models, and designed to cover incident response steps from triaging new cloud incidents to generating mitigation plans for the SREs. %Once initialized, the ARCA can generate mitigation plans without human intervention and this allows SREs to spend much less resource on retrieving old issues and hence they can focus more on the newly emerged challenging issues. Importantly, we developed ARCA using using off-the-shelf machine learning models. 
The ARCA multimodal RAG search mechanism (Sec. 3) is an original contribution of our effort.  The future ARCA will expand these  data modes and enlarge ARCA's multimodal pattern-matching capabilities.

 To test the end-to-end effectiveness of ARCA, we  created a data set of 800 bug reports collected from micro service systems in a controlled environment. The bug reports are typical Bugzilla incident reports of the kind users employ to request issue resolution. Each contains three components: 1) the user's incident description; 2) a log file collected from the docker container of the faulty service and 3) a time sequence of performance metrics collected from the same container during the the fault. Although the bugs have very different features, all trace to root causes associated with three widely recognized cloud computing issues: computations that exceeded time limits, memory leaks and network delays. In the evaluation, ARCA achieves 92\% accuracy in triage and 72\% accuracy in finding the correct mitigation plan. We have also tested the efficacy of individual components in ARCA using established data sets. %The evaluation methodology and results will be disclosed in Section 4, after which we discuss ideas for future work in Section 5.

\section{Related Work}
Before we dive into  details we review related work that shapes our thinking.

\subsection{Retrieval Augmented Generation}
The RAG paradigm is in widespread use %Content generation using RAG LLM is a prominent paradigm that combines the strengths of information retrieval and generative modeling to enhance the accuracy and contextual relevance of text generation tasks 
\cite{fb-rag, gao2024-rag-survey}. In this approach, a query is first transformed into a vector representation and an approximate nearest neighbor search is then used to fetch relevant documents from a knowledge base.  The retrieved content is then provided as auxiliary input to a generative model, typically an LLM. This extra ``context'' allows the model to ground its outputs in factual, up-to-date, or domain-specific information, reducing hallucinations and offering a way to continuously update the knowledge base without retraining models. RAG is effective for question answering \cite{karpukhin-etal-2020-dense}, summarization \cite{laban2024summaryhaystackchallengelongcontext}, and code generation \cite{parvez2021retrievalaugmentedcodegeneration}, and has been shown to significantly improve LLM performance and interpretability. Prior work on multimodal RAG has focused on the visual domain (text used to describe images).  %Moreover, the modular nature of RAG facilitates adaptability across domains by allowing updates to the retrieval database without retraining the generative model, making it a scalable and efficient solution for knowledge-intensive tasks. 
In ARCA, however, we need a RAG system specially for IT-Ops/AI-Ops. To the best of our knowledge, our work is the first to explore this form of multimodality.

\subsection{Prompting and Reasoning}
Prompt engineering is central to RAG LLM design. %Prompting has emerged as another powerful paradigm in natural language processing, enabling LLMs to perform diverse tasks with minimal training. 
One prompting technique, few-shot learning \cite{brown2020fewshot}, %, a key technique in this paradigm, uses carefully designed prompts containing a small number of task-specific examples to guide the model's behavior.
leverages the in-context learning capabilities of LLMs, guiding models from structure and examples in the prompt (without updates to model weights). A second, Chain of Thought \cite{wei2023cot}, takes a further step by structuring the prompt in a way that encourages step-by-step reasoning.  This has been shown to improve LLM performance on tasks requiring multi-step logical inference, arithmetic, or complex decision-making. In combination these two techniques achieve state-of-the-art performance across various domains including mathematics, common-sense reasoning, and question answering.  We adopt both in ARCA.

\subsection{AI-Ops}
ARCA is also inspired by prior work in AI-Ops \cite{AI-Ops-Survey}, notably for processing logs and telemetric data. LogCluster \cite{LogCluster} introduced techniques for clustering log records to assist in bug detection using a weighted encoding, and subsequent work used LLMs to summarize abnormalities in logs \cite{unilog, log_abnormal}. We used labeled log records from one of these efforts, LogHub\cite{LogHub}, for our evaluation.

We noted our interest in combining application instrumentation with text records from logs.  Prior studies have explored aspects of this question, notably by using deep neural networks for anomaly detection in multivariant time-series data. For example, Microsoft has proposed an anomaly detector based on Convolutional Neural Network (CNN) \cite{MSFT_Anomaly}, while Alibaba describes an encoder-decoder architecture in RobustTAD \cite{RobustTAD} and Tencent used a VAE network \cite{Tencent-VAE} for the same purpose.

Detecting anomalies in cloud platforms using telemetric performance data requires handling potentially noisy high-dimensionaldata. Li et al. (2024) have explored this problem and proposed a methodology for noise-tolerant self-
supervised learning \cite{RSDT} that combines tensor decomposition with self-supervised learning to capture relevant features and identify anomalies in time series data. For tabular data, the anomaly detection technique described in \cite{ADTD} shows that LLMs can detect anomalies by converting data into text and directing the models to find outliers. That effort went on to optimize performance by fine-tuning open-source LLMs using synthetic data.
In contrast, existing AI-Ops tools (including those we cited) have generally been limited to a single data modality. 

\section{How does ARCA work?}
\begin{figure*}[ht]
  \includegraphics[width=.8\textwidth,height=1.9in]{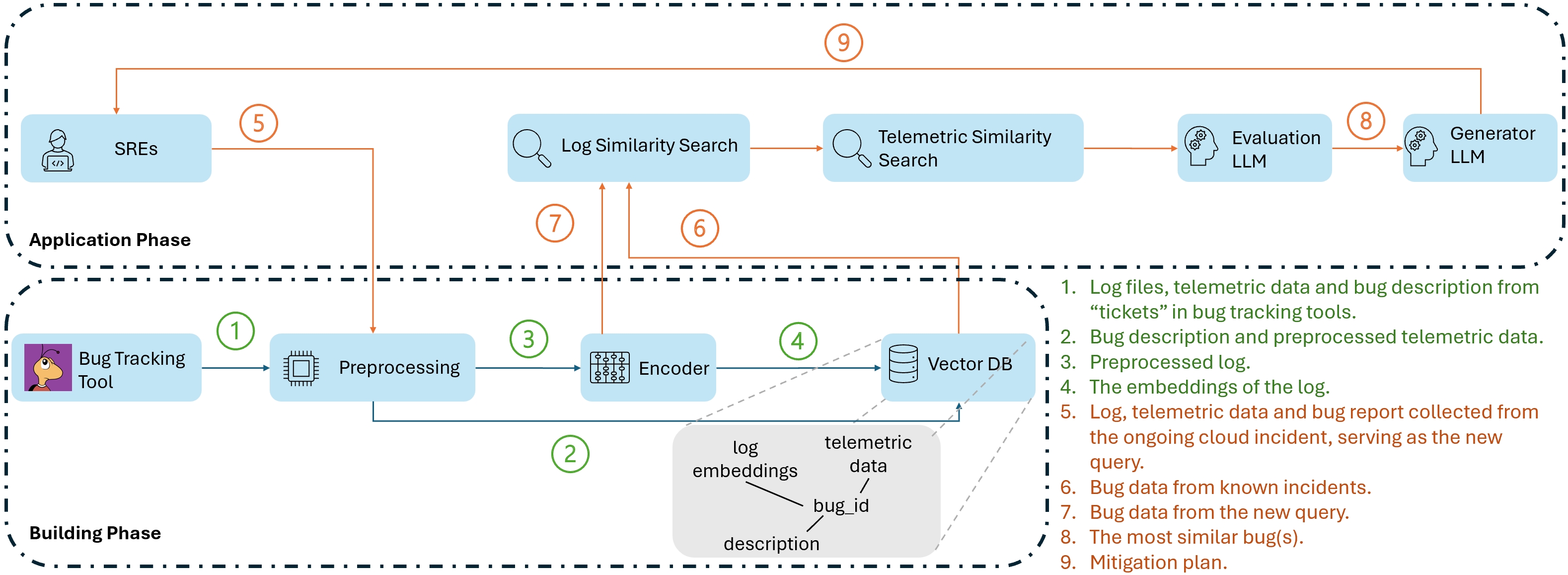}
  \caption{ARCA workflow in its \textcolor{green}{building} and \textcolor{brown}{query} phases.}
  \Description{ARCA workflow in the building phase and query phases.}
  \label{fig:system}
\end{figure*}

ARCA runs in two phases (Fig.~\ref{fig:system}): building the multimodal knowledge base of historical bugs and then querying it. Below we focus on a bug tracking use case, but the idea generalizes to other incident-analysis scenarios.
\subsection{Building Phase}
To deploy ARCA, we first collect and process data from existing solved bugs retrieved from bug tacking tools and then use the collected data to form a knowledge base. After creating the knowledge base, users can query ARCA for new and ongoing incidents, and the system will automatically generate a mitigation plan for each SRE.

\subsubsection{Data Sources}
We assume that software incidents are reported through tickets in a bug-tracking system such as Bugzilla. Each bug ticket contains multiple data modalities, e.g., bug descriptions (natural language), performance metrics (time sequences of numerical multi-variant data), logs (semi-structured machine-generated event reports), etc. In ARCA, we strive to find a mitigation plan by reasoning across the different modes of data.

A bug tracking system works like an online bulletin board similar to Reddit. Progress towards resolving bugs is tracked as follow-up posts to the original post initiated by the staff member who found the incident. To collect data to form a knowledge base, we keep our attention to the following steps within the life cycle of a bug ticket: (1) The first post, which includes a textual description of the problem. (2) The ticket assignment post, which reflects the judgment of a human triage specialists and has a fixed format. (3) Data collection posts with attachments: these are often data collected by the SRE team using tools they found relevant and is the step at which ARCA can learn from data modalities other than natural language. (4) The last post: the last post of a closed ticket is usually the diagnosis of the issue and the following mitigation. Notice that each category of posts and data hints at a its own similarity metric: rather than a single metric for all types of data, we need a unified metric spanning multiple modalities and robust against missing data (some reports may cite data that other related reports omit).

\subsubsection{Build A Multi-modal Knowledge Base} With the logs, performance metrics and bug descriptions retrieved from the prior step, we can build a knowledge base that associates related information. The idea behind the knowledge base is to do up front work so that later, we can quickly find similar bugs by comparing their logs and telemetric data during a triage step and then rapidly retrieve the corresponding bug descriptions to help create a mitigation plan.

An important design choice in ARCA is to use {\em search augmentation} to improve answer quality instead of storing the potential answers in the LLM parameters via a technique such as fine-tuning or training "Expert Heads" in a Mixture of Experts (MoE) models. We adopted search augmentation because creating a knowledge base, similar to creating a database, is much less resource-intensive than LLM training. Search augmentation additionally leaves us the freedom to update the knowledge base as information evolves; in contrast, pretrained models are rigid and hence unable to learn dynamically without some form of fine-tuning (which would often require resources on the same scale as were used during the original model tuning procedure). One drawback this decision is that it can increase latency, but we will show that ARCA is rapidly responsive and fully suited to interactive exploration of puzzling anomalies by SRE teams.

To enable fast similarity search among logs instead of directly searching the text space of the logs, ARCA maps ({\em embeds}  processed log snippets to a high-dimensional latent space: the embedding space.  The system will later use cosine similarity to quantify the difference between two log snippets. Calculating cosine similarity only involves calculating the product of two matrices, which can be carried out at very high speeds, particularly with the help of a GPU.  The task is much quicker than searching in the text space.  To further accelerate the similarity search on very large data sets, ARCA uses approximate K-nearest neighbors to organize the log embeddings in two tiers. To find the most similar log embeddings, we first look for the closest centroids.  The assumption is that these event clusters will contain the embeddings most relevant to the incident report.  We then use cosine similarity again, but now include performance metrics in our approximate similarity test. To enable this we first convert the performance metrics to a vector during our log preprocessing step by aligning the telemetric data of variant lengths and sources. Then, we store the vector in the knowledge base and via the bug id, can we associate it with other pieces of information collected from the same bug.

ARCA keeps  bug descriptions in natural language because they may contain important details that stood out to the human observer of the issue and hence are likely to be of high value to the tasks performed out by the Evaluation LLM in later steps. Additionally, bug resolution descriptions contain mitigation plans which proved effective in the past, and the Generator LLM can use those to propose a new plan to mitigate the ongoing issues.

ARCA's embedding space contains 3072-dimensional vectors of 32-bit floating point numbers, and we embed the logs using the "text-embedding-3-large" model from OpenAI. The preprocessed performance metrics are represented as 21-dimensional vectors of 32-bit floating point numbers. The knowledge base in ARCA is made of 3 object stores, with one for each of the log embedding, vectorized performance metrics and bug descriptions. We also maintain a mapping relationship between them. In the future we plan to allow dynamic additions to the database, but the PoC works with a static data set. 

\subsubsection{Process Log Files}
ARCA supports two data modalities: logs of semi-structured text and bug descriptions containing human inputs in natural language. We preprocess the logs prior to improve the accuracy of ARCA's triage technique.

The logs we consider are created by a variety of applications and systems services, and take the form of text files in which system maintenance messages, warnings and errors, anomaly notations, and other reporting can be intermixed. After examining the log files in our evaluation data set, we have found that

\begin{enumerate}
    \item A relatively small subset of log lines are relevant to any given incident.
    \item Log records of a given type are formatted in similar ways. For example, heartbeat messages for the same component only differ in their timestamps.
    \item For any single incident, a log may contain multiple relevant data modalities: text, tabular data, time-series data, etc.
\end{enumerate}

ARCA filters log contents by retrieving the bugs that show a "similar pattern" in logs in the query step. To keep the LLM focused on important features, it is important that the log contents visible to the LLM be relevant to the issues flagged in the problem report, and free of irrelevant information if that information might be of value for maintaining the system or other purposes.  Accordingly, we run a Feature Extraction LLM that we configure to remove repetitious content and extract  data that distinguishes each record from the others occuring at the same time, like the error messages, special events, performance metric readings, etc. We additionally convert all the data modalities that we encounter to text.  Length considerations precluded reproducing the prompt here, but we do include it as Appendix A, Fig. \ref{fig:log_prompt}.

\begin{figure}[t]
  \centerline{\includegraphics[width=\linewidth]{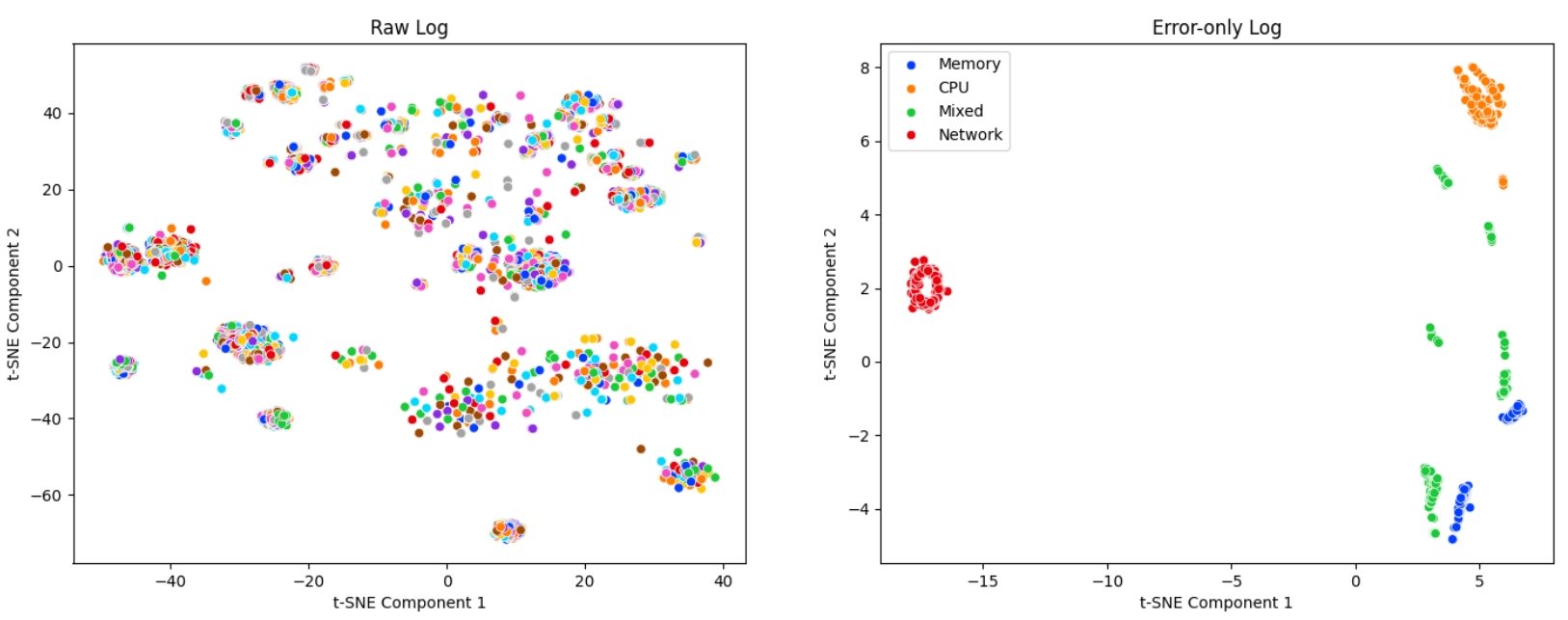}}
  \caption{t-SNE (t-distributed stochastic neighbor embedding) of the embedded log content. The x- and y-axes show the coordinates in the t-SNE embedding space.}
  \label{fig:tsne}
\end{figure}

To assess the efficacy of this log preparation approach we  processed 800 log files from our data set using OpenAI's gpt-4o as the Feature Extraction LLM. First, we generated embeddings from the raw log content with no preprocessing. Then, we preprocessed the logs and embedded only the filtered and aggregated outputs of the Feature Extraction LLM. We used t-SNE \cite{faiss} to project the high dimensional embedding space to a 2-D image while maintaining relative Euclidian distances. Doing so yielded the images seen in Fig. \ref{fig:tsne}, where each dot represents an embedding. As we can see, the embedding of processed logs (the right picture) resolves more clearly, showing a cleaner clustering pattern with far fewer clusters than for the raw log (the left picture): evidence that this step achieved its goals. We additionally colored the dots in both images to signify  root cause labels. As we can easily see, the dots from the same root cause, i.e., memory, CPU and network, are correctly clustered after preprocessing but were jumbled before doing so. Especially interesting are the green dots, for incidents in which a mix of CPU and memory issues simultaneously caused degraded system performance.  These green data points are correctly located between the clusters for CPU issues and those for memory issues.

\subsubsection{Align Telemetric Data}
To enable a similarity search, it is necessary to convert telemetry data to a fixed length vector. Raw  data can be highly platform-specific: a matrix with one row per time stamp and a column for each performance counter (CPU utilization, memory utilization, etc), but potentially with missing data due to faults and timeouts, idiosyncratic formats and units, and including hardware-specific metrics.  
%
%It's common to see the matrix come with different number of rows as data collection periods vary. The telemetry matrix can additionally have missing data points, typically because a service is unresponsive at the moment the log record is created. It's also common to see data collected by different software components reported using different performance counters even if the underlying property is the same. Further, the absolute value of performance counter readings can be highly sensitive to the underlying hardware and the version of the running software.
%
    To overcome these issues, ARCA focuses on a set of 7 docker performance counters, all of which are commonly available when diagnosing cloud microservice incidents. These track CPU and memory utilization, network I/O, block device I/Os, average operation latency, and socket errors. Servers are highly heterogeneous, hence raw values are not directly comparable.  Accordingly, we calculate the normalized first order gradient, the average value and the standard deviation for each time series. In this way, we can convert the matrix of performance counter readings to a vector of 21 floating point numbers.

\subsection{ARCA-PoC Phases}

ARCA-PoC runs in two sub-phases: the query phase and the generating phase. In the query phase, we interrogate the populated knowledge base by carrying out the similarity search on log embeddings and vectorized performance metrics. This phase is analogous to the triage step and the output are the textual descriptions of similar bugs. The descriptions are then sent to the generating phase to create a mitigation plan for the SREs.

\subsubsection{Query Phase}

Once our knowledge base has been populated, ARCA performs an approximate match query using posts associated with a new incident as its query prompts. The methodology used to extract the relevant aspects of the incident is quite similar to the one used to build the knowledge base, and yields an embedding vector that we can understand as an abstract representation of the new incident in the knowledge space.  Our goal is to perform an approximate nearest neighbor (ANN) search.  We do this in two steps: first, we identify cluster centroids closest to the query embedding, and then within those clusters perform a search for known prior incidents with similar characteristics.  Here, ARCA departs slightly from common RAG approaches that only retrieve the top tens of documents based on the similarity score. Instead, ARCA retrieves the top hundreds of bugs as reported from the similarity search. This is because ARCA treats similarity search as a triage step for the purpose of  coarsely categorizing a bug by placing it within a family of issues so that corresponding SREs can chime in. For example, if a bug seems to be CPU-related, it could be assigned to SREs working on performance issues, ones working on scheduling, and ones investigating disruptions associated with locking. With just a small number of approximate matches we might miss some relevant categories, but with hundreds of approximate machines, we have a high likelihood of routing the issue to all SREs that might have insight into the issue.

From the bugs with similar log patterns, we additionally perform a second-round KNN search in the high-dimensional space of the vectorized performance metrics. Here, an issue of cost arises: our work uses OpenAI language-generation APIs that are billed on a per-use basis.  Accordingly, we only use one tenth of our prior report candidates for generation of the bug explanation hypotheses that the developer will be shown. In the evaluation, we will show that this step of filtering will not hurt the overall accuracy.

In ARCA, we use FAISS library \cite{tsne} to carry out the similarity search so that it will run on GPU accelerators. We have tried to retrieve from top 100 bugs to top 500 bugs and we can reach a triage success rate as high as 92\%. We will discuss the effect brought by different number of retrieved bugs in evaluation section.

\subsubsection{Generating Phase}
In the generating phase, we first use an Evaluation LLM to find the bug whose description most closely fits each incident. We pass the description of the bug fix (which  contains the mitigation plan) to the Generator LLM, which in turn produces text explaining the choice and suggesting a new mitigation plan to the SREs. The approach is similar to a concept sometimes referred to as {\em LLM-as-a-judge} \cite{llm-as-judge} (the corresponding prompt details are included in Appendix A, Fig. \ref{fig:evaluation_llm}). To improve accuracy, we employ a Chain of Thought prompting style (Appendix A, Fig. \ref{fig:evaluation_cot}), using a series of  similar CoT prompts in accordance with standard practice in few-shot learning.  The output of this step is the closest resolved bug. We then prompt the Generator LLM with the input shown in Fig. \ref{fig:generate}. An additional benefit brought by using LLM-as-a-judge is that we can ask the evaluating LLM to explain how and why it reached certain conclusions, either in summary form or even as a sequence of step-by-step decisions, allowing the SREs to better understand the results and hence increasing confidence in its coverage.  Were ARCA to operate in a single step, it would have more of a black-box feel that SREs might distrust.

Our design is human-centric:  ARCA generates mitigation plans and reports them  to SREs for final review, together with illustrative data drawn from any similar incidents it found. We are not considering direct intervention by ARCA at this time, in part because some privileged commands (like restarting critical services) require privilage escalation and should not occur without close scrutiny and Dev-Ops (human) approval. A benefit is that ARCAs ability to identify similar prior incidents may be helpful to SREs even if its proposed mitigation plans are flawed. To obtain recommendations with a natural tone and style, ARCA uses gpt-4o for both the evaluating and generation LLM stages.

\section{Experimental Results}
To evaluate our work, we first build a data set for 800 bug tickets containing descriptions, logs and performance metrics. Then we build ARCA's knowledge base using 700 bug tickets, saving 100 to use when testing the PoC's performance.

\subsection{Data Set}
Our data set of bugs arising in micro service systems is typical of modern cloud infrastructures. To keep our data set as general as possible, we keep our attention only to the bug features reported from the docker container, including the docker logs and the performance metric readings from the "top" command, without any application-level features.

We use a micro-service workload generator,  "DeathStar" \cite{deathstar} to run different micro-service applications, like "HotelReservation", "SocialNetwork", etc. As the application executes, we inject errors. To load the CPU, we modified the benchmark so before processing a new request, the application performs a CPU-intensive operation. We also increase the number of requests per second during runtime until the application crashes from overload. To simulate a memory leak we modify the benchmark by introducing a memory allocation in the call back function but intentionally not freeing the memory. Finally, to increase network delays, we introduce a random sleep in the call back function. To make the challenge harder, we have introduced a fourth category of error that causes both a memory leak and a long-running computation, resulting in two possible crash types.

Each of the four categories of injected errors are used to create 100 experiments, which we diversify by tweaking settings. We run each experiment twice so that we can use the data set we can automatically label an experiment run with its closest bug, which is the run generated from the same experiment configuration, yielding 4*100*2=800 bug incidents.  For each bug, we use gpt-4o to generate a human readable bug report. In the generating prompt, we have provided the root causes like "the issue is caused by a random delay in every invocation of the call back function X" to ensure that the bug report contains meaningful mitigation plans. We have also instructed the LLM to decribe the bug by summarizing the performance metric readings and the logs.  We thus obtain 800 bug tickets that contain the bug descriptions with mitigation plans, the time-series of performance metrics and the logs.

To evaluate the efficacy of our similarity search in the log embedding space, which is the key of the RAG system, we use public data sets from four supercomputing systems: BGL, Thunderbird, Liberty, and Spirit \cite{HPC4}.

\subsection{End-to-end Evaluation}
\begin{figure}[t]
  \centerline{\includegraphics[width=.9\linewidth]{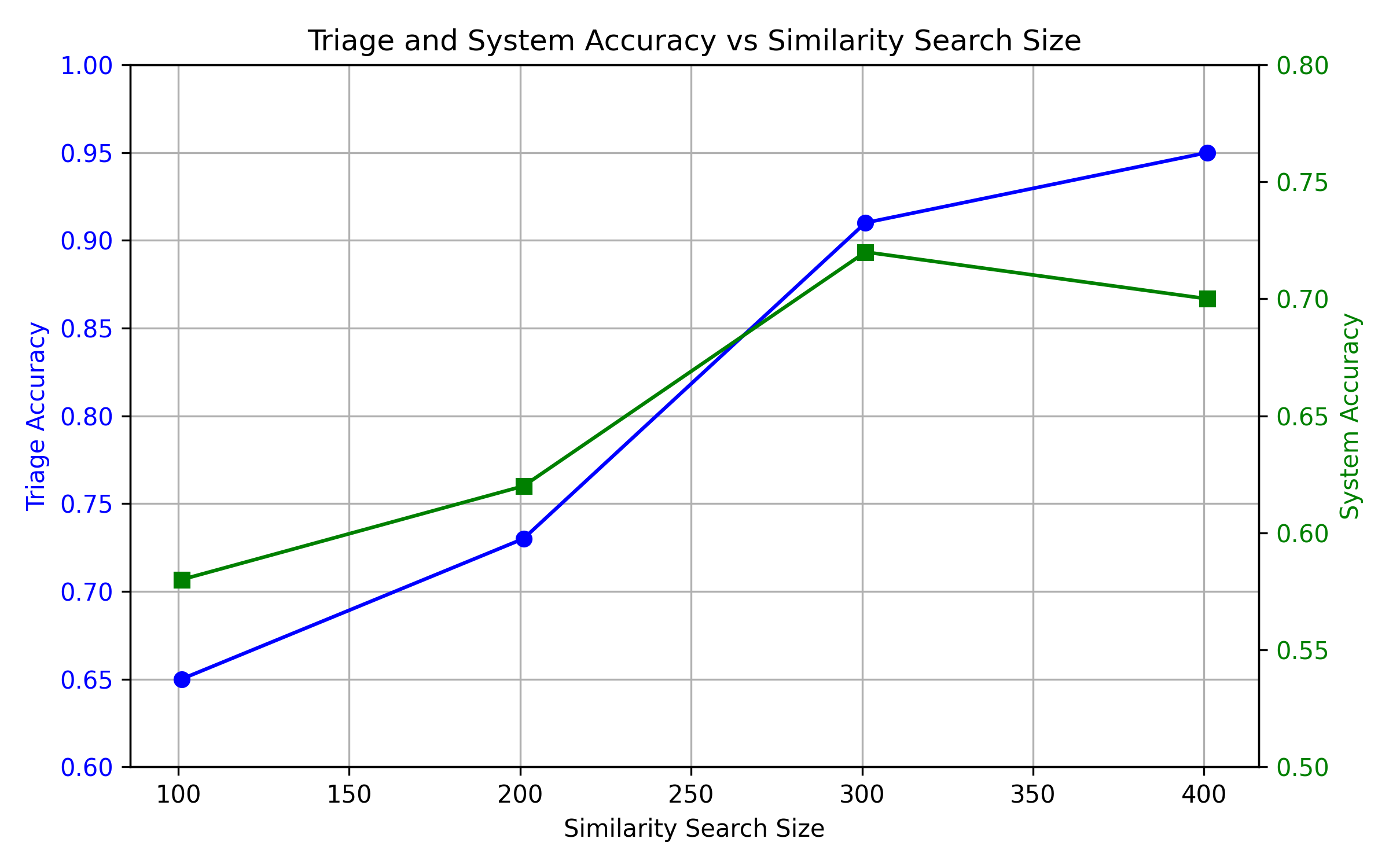}}
  \caption{Accuracy of ARCA-PoC. The x-axis represents the output size of the similarity search. The left y-axis shows the triage accuracy, while the right y-axis shows the system accuracy.}
  \label{fig:arcr_acc}
\end{figure}

We first study the effect of using different numbers of nearest neighbors reported from the similarity search module. This is also the size of the output from the triage step. So we compare both the triage accuracy and the system accuracy. For a triage operation to be accurate, ARCA needs to include the closest bug in the output of the triage steps. For the whole system to be accurate, ARCA needs to pick the labeled closest bug as the output of the Evaluation LLM. The results are shown in Fig. \ref{fig:arcr_acc}. To account for the randomness introduced by the LLMs, we evaluated the average performance on 300 queries for each setting and for each query, we repeated the experiment for 3 times. 
% The results are shown in Fig. \ref{fig:arcr_acc} and \ref{fig:arcr_cost}
%we  repeated each test 300 times and report the average values in 

In our test, we increase the log similarity search output size from 100 to 400, and we filter out 20\% of the chosen bugs in the similarity search using telemetric data. As we can see, triage accuracy increases steadily with the raise of the triage set size. However, the overall system accuracy drops when we increase the similarity search size from 300 to 400. Upon inspection we found that when the similarity search size is small (less than 200), the right answer is not presented in the input prompt. This ceases to be an issue with larger set sizes. Interestingly, however, although triage accuracy at set size 400 is significantly higher than that for size 300, overall system accuracy drops: the Evaluation LLM apparently becomes overwhelmed by choices.

\begin{figure}[t]
  \centerline{\includegraphics[width=.9\linewidth]{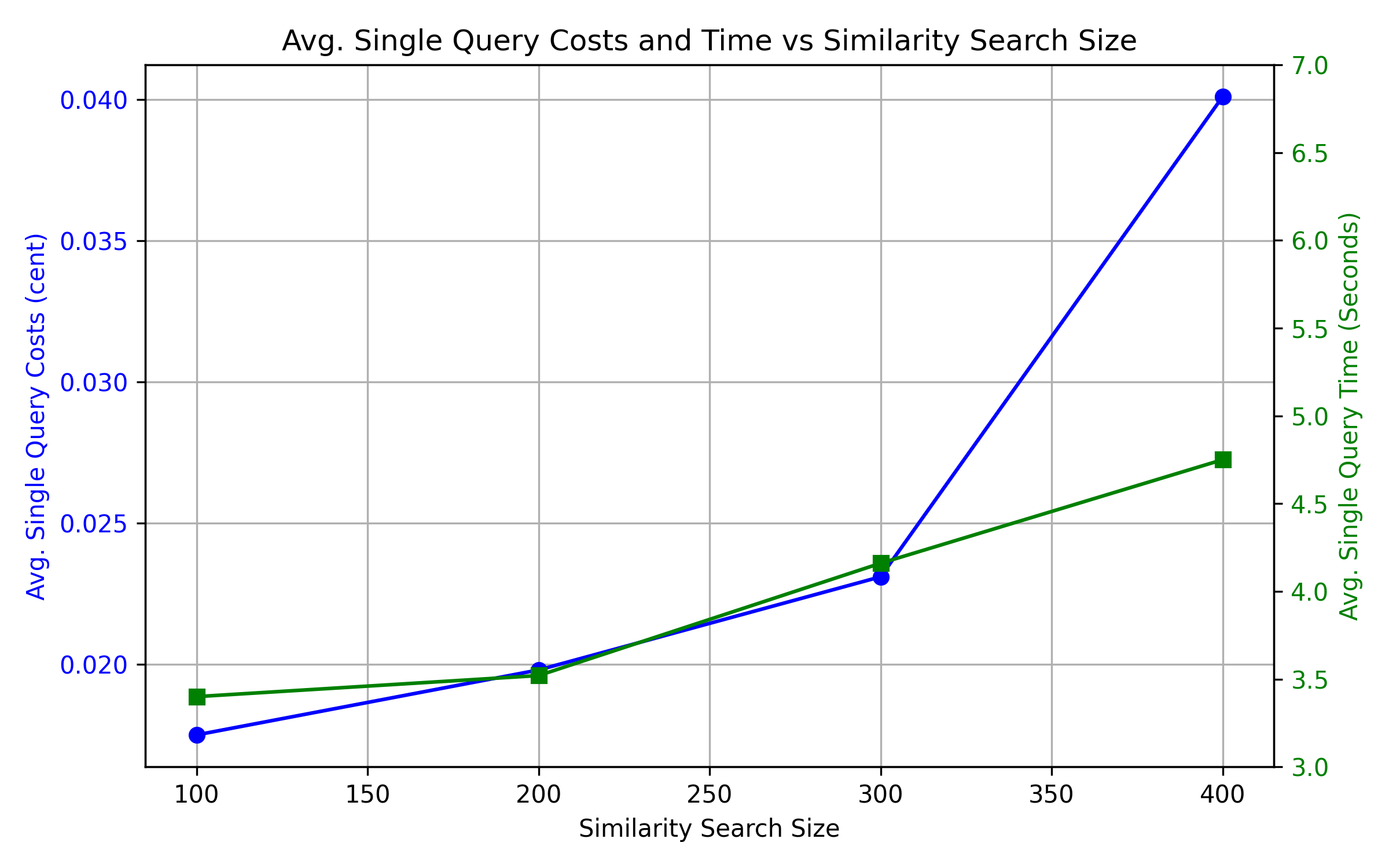}}
  \caption{Cost analysis of using ARCA-PoC. The x-axis represents the output size of the similarity search. The left y-axis shows the average cost of a single query in US cents, while the right y-axis shows its average time consumption.}
  \label{fig:arcr_cost}
\end{figure}
It's also worth noting that we cannot increase the similarity search output size without limit. GPT-4o, the LLM we use for our Evaluation LLM, has a context window size limit of 30,000 tokens and the input cannot be longer than that. This token window limit corresponds to a triage output set size of slightly more than 400.

We also evaluated time and financial cost per query. Gpt-4o uses a decoder-only neural network structure and hence the longer the input in tokens, the more time it will take to generate an answer. Also, OpenAI charges clients on the basis of the number of tokens computed. Taking all these considerations together, we arrive at the results shown in Fig. \ref{fig:arcr_cost}. As we can see, for large group size,  generation is significantly slower and cost mounts substantially.

\subsection{Evaluation of Similarity Search}
There are two similarity search steps in ARCA, with one in the high-dimensional space of log embedding and the other in telemetric encoding. We have evaluated the efficacy of each  step against the combined performance, and reported the results in Table. \ref{tab:modality_comparison}, where we pick the triage size to be 300. The key finding is that multi-modal similarity search  saves time and money relative to approaches that require one by one searches across modes followed by a human integrative activity.

\begin{table}[h]
    \centering
    \begin{tabular}{c|c|c|c}
        \hline
        Data Modes & Accuracy & Cost (cents) & Time (s) \\
        \hline
        Telemetric Data Only & 0.34 & 2.81 & 4.67 \\
        Log Only & 0.72 & \textbf{2.31} & \textbf{4.16} \\
        Telemetric Data + Log & \textbf{0.74} & 2.89 & 4.89\\
        \hline
    \end{tabular}
    \caption{Comparison of the efficacy via utilizing different modes of data.}
    \label{tab:modality_comparison}
\end{table}

% \begin{figure}[t]
%   \centerline{\includegraphics[width=.9\linewidth]{images/bar.png}}
%   \caption{Evaluation of each similarity search step.}
%   \label{fig:bar}
% \end{figure}

\subsection{ARCA as A Log Clustering Tool}
Very similar to log clustering tools, ARCA's RAG-LLM based log processing module can be used alone to detect anomalies in logs. In our evaluation, we use public log data sets reported from 4 supercomputing labs and report the results in Table. \ref{tab:log}. The numbers before '/' are from ARCA and the one after are the state-of-the-art numbers reported in \cite{log_abnormal, logformer}, which are achieved through proprietarily fine-tuned LLMs. %(* for lacking recent publicly reported SOTA performance.) 
From the results, ARCA-PoC outperforms on all data sets despite requiring only off-the-shelf embedding LLMs.

\begin{table}[h]
    \centering
    \begin{tabular}{c|c|c c}
        \hline
        Data Set & F1-Score & Recall & Precision \\
        \hline
        BGL & \textbf{0.995}/0.976 & \textbf{0.99}/0.982 & \textbf{1}/0.970 \\
        Thunderbird& \textbf{0.984}/0.97 & 0.975/\textbf{0.99} & \textbf{1}/0.97 \\
        Spirit & \textbf{0.993}/0.992 & 0.986/\textbf{0.999} & \textbf{1}/0.984\\
        Liberty & 0.986/* & 0.986/* & 0.986/*\\
        \hline
    \end{tabular}
    \caption{Evaluation of using ARCA-PoC as a log clustering tool. *:  For the Liberty data set, public baseline data is not available.}
    \label{tab:log}
\end{table}

\section{Conclusions and Future Work}
ARCA is a work in progress, but already confirms the promise of the multimodal RAG LLM approach to searching the complicated incident report databases that arise when troubleshooting cloud-hosted applications. In work still underway, we are  investigating other possible ways to organize the ARCA knowledge base, including the option of using similarity search algorithms beyond the form of cosine similarity used in  the ANN step. We expect that this will be needed as we expand the data modality coverage of the ARCA platform to include performance metrics and traces. Synthesis of generated answers that incorporate observations from multiple modalities raises especially interesting questions for study.

\begin{acks}
We would like to thank Tiancheng Yuan for his insight on RAG LLMs and Miles Bramwit for his efforts on data collection. We are also grateful for support we received from Siemens, NVIDIA, Cisco and Microsoft.
\end{acks}

\bibliographystyle{ACM-Reference-Format}
\bibliography{arca}


\begin{thebibliography}{24}

%%% the Web address.

\ifx \showCODEN    \undefined \def \showCODEN     #1{\unskip}     \fi
\ifx \showDOI      \undefined \def \showDOI       #1{#1}\fi
\ifx \showISBNx    \undefined \def \showISBNx     #1{\unskip}     \fi
\ifx \showISBNxiii \undefined \def \showISBNxiii  #1{\unskip}     \fi
\ifx \showISSN     \undefined \def \showISSN      #1{\unskip}     \fi
\ifx \showLCCN     \undefined \def \showLCCN      #1{\unskip}     \fi
\ifx \shownote     \undefined \def \shownote      #1{#1}          \fi
\ifx \showarticletitle \undefined \def \showarticletitle #1{#1}   \fi
\ifx \showURL      \undefined \def \showURL       {\relax}        \fi
% invisible to TeX
\providecommand\bibfield[2]{#2}
\providecommand\bibinfo[2]{#2}
\providecommand\natexlab[1]{#1}
\providecommand\showeprint[2][]{arXiv:#2}

\bibitem[Brown et~al\mbox{.}(2020)]%
        {brown2020fewshot}
\bibfield{author}{\bibinfo{person}{Tom~B. Brown}, \bibinfo{person}{Benjamin Mann}, \bibinfo{person}{Nick Ryder}, \bibinfo{person}{Melanie Subbiah}, \bibinfo{person}{Jared Kaplan}, \bibinfo{person}{Prafulla Dhariwal}, {and} \bibinfo{person}{et al.}} \bibinfo{year}{2020}\natexlab{}.
\newblock \bibinfo{title}{Language Models are Few-Shot Learners}.
\newblock
\newblock
\showeprint[arxiv]{2005.14165}~[cs.CL]
\urldef\tempurl%
\url{https://arxiv.org/abs/2005.14165}
\showURL{%
\tempurl}


\bibitem[Cheng et~al\mbox{.}(2023)]%
        {AI-Ops-Survey}
\bibfield{author}{\bibinfo{person}{Qian Cheng}, \bibinfo{person}{Doyen Sahoo}, \bibinfo{person}{Amrita Saha}, \bibinfo{person}{Wenzhuo Yang}, \bibinfo{person}{Chenghao Liu}, \bibinfo{person}{Gerald Woo}, \bibinfo{person}{Manpreet Singh}, \bibinfo{person}{Silvio Saverese}, {and} \bibinfo{person}{Steven C.~H. Hoi}.} \bibinfo{year}{2023}\natexlab{}.
\newblock \bibinfo{title}{AI for IT Operations (AIOps) on Cloud Platforms: Reviews, Opportunities and Challenges}.
\newblock
\newblock
\showeprint[arxiv]{2304.04661}~[cs.LG]
\urldef\tempurl%
\url{https://arxiv.org/abs/2304.04661}
\showURL{%
\tempurl}


\bibitem[Gan et~al\mbox{.}(2019)]%
        {deathstar}
\bibfield{author}{\bibinfo{person}{Yu Gan}, \bibinfo{person}{Yanqi Zhang}, \bibinfo{person}{Dailun Cheng}, \bibinfo{person}{Ankitha Shetty}, \bibinfo{person}{Priyal Rathi}, \bibinfo{person}{Christina Delimitrou}, {and} \bibinfo{person}{et al.}} \bibinfo{year}{2019}\natexlab{}.
\newblock \showarticletitle{An Open-Source Benchmark Suite for Microservices and Their Hardware-Software Implications for Cloud \& Edge Systems}. In \bibinfo{booktitle}{\emph{Proceedings of the Twenty-Fourth International Conference on Architectural Support for Programming Languages and Operating Systems}} (Providence, RI, USA) \emph{(\bibinfo{series}{ASPLOS '19})}. \bibinfo{publisher}{Association for Computing Machinery}, \bibinfo{address}{New York, NY, USA}, \bibinfo{pages}{3–18}.
\newblock
\showISBNx{9781450362405}
\urldef\tempurl%
\url{https://doi.org/10.1145/3297858.3304013}
\showDOI{\tempurl}


\bibitem[Gao et~al\mbox{.}(2021)]%
        {RobustTAD}
\bibfield{author}{\bibinfo{person}{Jingkun Gao}, \bibinfo{person}{Xiaomin Song}, \bibinfo{person}{Qingsong Wen}, \bibinfo{person}{Pichao Wang}, \bibinfo{person}{Liang Sun}, {and} \bibinfo{person}{Huan Xu}.} \bibinfo{year}{2021}\natexlab{}.
\newblock \bibinfo{title}{RobustTAD: Robust Time Series Anomaly Detection via Decomposition and Convolutional Neural Networks}.
\newblock
\newblock
\showeprint[arxiv]{2002.09545}~[cs.LG]
\urldef\tempurl%
\url{https://arxiv.org/abs/2002.09545}
\showURL{%
\tempurl}


\bibitem[Gao et~al\mbox{.}(2024)]%
        {gao2024-rag-survey}
\bibfield{author}{\bibinfo{person}{Yunfan Gao}, \bibinfo{person}{Yun Xiong}, \bibinfo{person}{Xinyu Gao}, \bibinfo{person}{Kangxiang Jia}, \bibinfo{person}{Jinliu Pan}, \bibinfo{person}{Yuxi Bi}, {and} \bibinfo{person}{et al.}} \bibinfo{year}{2024}\natexlab{}.
\newblock \bibinfo{title}{Retrieval-Augmented Generation for Large Language Models: A Survey}.
\newblock
\newblock
\showeprint[arxiv]{2312.10997}~[cs.CL]
\urldef\tempurl%
\url{https://arxiv.org/abs/2312.10997}
\showURL{%
\tempurl}


\bibitem[Guo et~al\mbox{.}(2024)]%
        {logformer}
\bibfield{author}{\bibinfo{person}{Hongcheng Guo}, \bibinfo{person}{Jian Yang}, \bibinfo{person}{Jiaheng Liu}, \bibinfo{person}{Jiaqi Bai}, \bibinfo{person}{Boyang Wang}, \bibinfo{person}{Zhoujun Li}, \bibinfo{person}{Tieqiao Zheng}, \bibinfo{person}{Bo Zhang}, \bibinfo{person}{Junran peng}, {and} \bibinfo{person}{Qi Tian}.} \bibinfo{year}{2024}\natexlab{}.
\newblock \bibinfo{title}{LogFormer: A Pre-train and Tuning Pipeline for Log Anomaly Detection}.
\newblock
\newblock
\showeprint[arxiv]{2401.04749}~[cs.LG]
\urldef\tempurl%
\url{https://arxiv.org/abs/2401.04749}
\showURL{%
\tempurl}


\bibitem[Huang et~al\mbox{.}(2022)]%
        {Tencent-VAE}
\bibfield{author}{\bibinfo{person}{Tao Huang}, \bibinfo{person}{Pengfei Chen}, {and} \bibinfo{person}{Ruipeng Li}.} \bibinfo{year}{2022}\natexlab{}.
\newblock \showarticletitle{A Semi-Supervised VAE Based Active Anomaly Detection Framework in Multivariate Time Series for Online Systems} \emph{(\bibinfo{series}{WWW '22})}. \bibinfo{publisher}{Association for Computing Machinery}, \bibinfo{address}{New York, NY, USA}, \bibinfo{numpages}{10}~pages.
\newblock
\showISBNx{9781450390965}
\urldef\tempurl%
\url{https://doi.org/10.1145/3485447.3511984}
\showDOI{\tempurl}


\bibitem[Johnson et~al\mbox{.}(2017)]%
        {faiss}
\bibfield{author}{\bibinfo{person}{Jeff Johnson}, \bibinfo{person}{Matthijs Douze}, {and} \bibinfo{person}{Hervé Jégou}.} \bibinfo{year}{2017}\natexlab{}.
\newblock \bibinfo{title}{Billion-scale similarity search with GPUs}.
\newblock
\newblock
\showeprint[arxiv]{1702.08734}~[cs.CV]
\urldef\tempurl%
\url{https://arxiv.org/abs/1702.08734}
\showURL{%
\tempurl}


\bibitem[Karpukhin et~al\mbox{.}(2020)]%
        {karpukhin-etal-2020-dense}
\bibfield{author}{\bibinfo{person}{Vladimir Karpukhin}, \bibinfo{person}{Barlas Oğuz}, \bibinfo{person}{Sewon Min}, \bibinfo{person}{Patrick Lewis}, \bibinfo{person}{Ledell Wu}, \bibinfo{person}{Wen tau Yih}, {and} \bibinfo{person}{et al.}} \bibinfo{year}{2020}\natexlab{}.
\newblock \bibinfo{title}{Dense Passage Retrieval for Open-Domain Question Answering}.
\newblock
\newblock
\showeprint[arxiv]{2004.04906}~[cs.CL]
\urldef\tempurl%
\url{https://arxiv.org/abs/2004.04906}
\showURL{%
\tempurl}


\bibitem[Laban et~al\mbox{.}(2024)]%
        {laban2024summaryhaystackchallengelongcontext}
\bibfield{author}{\bibinfo{person}{Philippe Laban}, \bibinfo{person}{Alexander~R. Fabbri}, \bibinfo{person}{Caiming Xiong}, {and} \bibinfo{person}{Chien-Sheng Wu}.} \bibinfo{year}{2024}\natexlab{}.
\newblock \bibinfo{title}{Summary of a Haystack: A Challenge to Long-Context LLMs and RAG Systems}.
\newblock
\newblock
\showeprint[arxiv]{2407.01370}~[cs.CL]
\urldef\tempurl%
\url{https://arxiv.org/abs/2407.01370}
\showURL{%
\tempurl}


\bibitem[Lewis et~al\mbox{.}(2021)]%
        {fb-rag}
\bibfield{author}{\bibinfo{person}{Patrick Lewis}, \bibinfo{person}{Ethan Perez}, \bibinfo{person}{Aleksandra Piktus}, \bibinfo{person}{Fabio Petroni}, \bibinfo{person}{Vladimir Karpukhin}, \bibinfo{person}{Naman Goyal}, {and} \bibinfo{person}{et al.}} \bibinfo{year}{2021}\natexlab{}.
\newblock \bibinfo{title}{Retrieval-Augmented Generation for Knowledge-Intensive NLP Tasks}.
\newblock
\newblock
\showeprint[arxiv]{2005.11401}~[cs.CL]
\urldef\tempurl%
\url{https://arxiv.org/abs/2005.11401}
\showURL{%
\tempurl}


\bibitem[Li et~al\mbox{.}(2024)]%
        {ADTD}
\bibfield{author}{\bibinfo{person}{Aodong Li}, \bibinfo{person}{Yunhan Zhao}, \bibinfo{person}{Chen Qiu}, \bibinfo{person}{Marius Kloft}, \bibinfo{person}{Padhraic Smyth}, \bibinfo{person}{Maja Rudolph}, {and} \bibinfo{person}{Stephan Mandt}.} \bibinfo{year}{2024}\natexlab{}.
\newblock \bibinfo{title}{Anomaly Detection of Tabular Data Using LLMs}.
\newblock
\newblock
\showeprint[arxiv]{2406.16308}~[cs.LG]
\urldef\tempurl%
\url{https://arxiv.org/abs/2406.16308}
\showURL{%
\tempurl}


\bibitem[Lin et~al\mbox{.}(2016)]%
        {LogCluster}
\bibfield{author}{\bibinfo{person}{Qingwei Lin}, \bibinfo{person}{Hongyu Zhang}, \bibinfo{person}{Jian-Guang Lou}, \bibinfo{person}{Yu Zhang}, {and} \bibinfo{person}{Xuewei Chen}.} \bibinfo{year}{2016}\natexlab{}.
\newblock \showarticletitle{Log Clustering Based Problem Identification for Online Service Systems}. In \bibinfo{booktitle}{\emph{2016 IEEE/ACM 38th International Conference on Software Engineering Companion (ICSE-C)}}. \bibinfo{pages}{102--111}.
\newblock


\bibitem[Man~Li and Tsung(2024)]%
        {RSDT}
\bibfield{author}{\bibinfo{person}{Lijun~Sun Man~Li, Ziyue~Li} {and} \bibinfo{person}{Fugee Tsung}.} \bibinfo{year}{2024}\natexlab{}.
\newblock \showarticletitle{Robust Self-Supervised Deep Tensor Decomposition for Corrupted Time Series Classification}. In \bibinfo{booktitle}{\emph{Anomaly Detection with Foundation Models}}. \bibinfo{address}{Jeju, South Korea}.
\newblock
\urldef\tempurl%
\url{https://adfmw.github.io/ijcai24/index.html}
\showURL{%
\tempurl}


\bibitem[Oliner and Stearley(2007)]%
        {HPC4}
\bibfield{author}{\bibinfo{person}{Adam Oliner} {and} \bibinfo{person}{Jon Stearley}.} \bibinfo{year}{2007}\natexlab{}.
\newblock \showarticletitle{What Supercomputers Say: A Study of Five System Logs}. In \bibinfo{booktitle}{\emph{37th Annual IEEE/IFIP International Conference on Dependable Systems and Networks (DSN'07)}}. \bibinfo{pages}{575--584}.
\newblock
\urldef\tempurl%
\url{https://doi.org/10.1109/DSN.2007.103}
\showDOI{\tempurl}


\bibitem[Parvez et~al\mbox{.}(2021)]%
        {parvez2021retrievalaugmentedcodegeneration}
\bibfield{author}{\bibinfo{person}{Md~R. Parvez}, \bibinfo{person}{Wasi~U. Ahmad}, \bibinfo{person}{Saikat Chakraborty}, \bibinfo{person}{Baishakhi Ray}, {and} \bibinfo{person}{Kai-Wei Chang}.} \bibinfo{year}{2021}\natexlab{}.
\newblock \bibinfo{title}{Retrieval Augmented Code Generation and Summarization}.
\newblock
\newblock
\showeprint[arxiv]{2108.11601}~[cs.SE]
\urldef\tempurl%
\url{https://arxiv.org/abs/2108.11601}
\showURL{%
\tempurl}


\bibitem[Ren et~al\mbox{.}(2019)]%
        {MSFT_Anomaly}
\bibfield{author}{\bibinfo{person}{Hansheng Ren}, \bibinfo{person}{Bixiong Xu}, \bibinfo{person}{Yujing Wang}, \bibinfo{person}{Chao Yi}, \bibinfo{person}{Congrui Huang}, \bibinfo{person}{Xiaoyu Kou}, \bibinfo{person}{Tony Xing}, \bibinfo{person}{Mao Yang}, \bibinfo{person}{Jie Tong}, {and} \bibinfo{person}{Qi Zhang}.} \bibinfo{year}{2019}\natexlab{}.
\newblock \showarticletitle{Time-Series Anomaly Detection Service at Microsoft}. In \bibinfo{booktitle}{\emph{Proceedings of the 25th ACM SIGKDD International Conference on Knowledge Discovery \& Data Mining}} \emph{(\bibinfo{series}{KDD ’19})}. \bibinfo{publisher}{ACM}, \bibinfo{pages}{3009–3017}.
\newblock
\urldef\tempurl%
\url{https://doi.org/10.1145/3292500.3330680}
\showDOI{\tempurl}


\bibitem[Schroeder and Gibson(2007)]%
        {com2}
\bibfield{author}{\bibinfo{person}{Bianca Schroeder} {and} \bibinfo{person}{Garth~A. Gibson}.} \bibinfo{year}{2007}\natexlab{}.
\newblock \showarticletitle{Disk Failures in the Real World: What Does an {MTTF} of 1,000,000 Hours Mean to You?}. In \bibinfo{booktitle}{\emph{5th USENIX Conference on File and Storage Technologies (FAST 07)}}. \bibinfo{publisher}{USENIX Association}, \bibinfo{address}{San Jose, CA}.
\newblock
\urldef\tempurl%
\url{https://www.usenix.org/conference/fast-07/disk-failures-real-world-what-does-mttf-1000000-hours-mean-you}
\showURL{%
\tempurl}


\bibitem[v.~d. Maaten and Hinton(2008)]%
        {tsne}
\bibfield{author}{\bibinfo{person}{Laurens v.~d. Maaten} {and} \bibinfo{person}{Geoffrey Hinton}.} \bibinfo{year}{2008}\natexlab{}.
\newblock \showarticletitle{Visualizing Data using t-SNE}.
\newblock \bibinfo{journal}{\emph{Journal of Machine Learning Research}} \bibinfo{volume}{9}, \bibinfo{number}{86} (\bibinfo{year}{2008}), \bibinfo{pages}{2579--2605}.
\newblock
\urldef\tempurl%
\url{http://jmlr.org/papers/v9/vandermaaten08a.html}
\showURL{%
\tempurl}


\bibitem[Wang et~al\mbox{.}(2025)]%
        {log_abnormal}
\bibfield{author}{\bibinfo{person}{Yuqing Wang}, \bibinfo{person}{Mika~V. Mäntylä}, \bibinfo{person}{Jesse Nyyssölä}, \bibinfo{person}{Ke Ping}, {and} \bibinfo{person}{Liqiang Wang}.} \bibinfo{year}{2025}\natexlab{}.
\newblock \bibinfo{title}{Cross-System Software Log-based Anomaly Detection Using Meta-Learning}.
\newblock
\newblock
\showeprint[arxiv]{2412.15445}~[cs.SE]
\urldef\tempurl%
\url{https://arxiv.org/abs/2412.15445}
\showURL{%
\tempurl}


\bibitem[Wei et~al\mbox{.}(2023)]%
        {wei2023cot}
\bibfield{author}{\bibinfo{person}{Jason Wei}, \bibinfo{person}{Xuezhi Wang}, \bibinfo{person}{Dale Schuurmans}, \bibinfo{person}{Maarten Bosma}, \bibinfo{person}{Brian Ichter}, \bibinfo{person}{Denny Zhou}, {and} \bibinfo{person}{et al.}} \bibinfo{year}{2023}\natexlab{}.
\newblock \bibinfo{title}{Chain-of-Thought Prompting Elicits Reasoning in Large Language Models}.
\newblock
\newblock
\showeprint[arxiv]{2201.11903}~[cs.CL]
\urldef\tempurl%
\url{https://arxiv.org/abs/2201.11903}
\showURL{%
\tempurl}


\bibitem[Zheng et~al\mbox{.}(2024)]%
        {llm-as-judge}
\bibfield{author}{\bibinfo{person}{Lianmin Zheng}, \bibinfo{person}{Wei-Lin Chiang}, \bibinfo{person}{Ying Sheng}, \bibinfo{person}{Siyuan Zhuang}, \bibinfo{person}{Zhanghao Wu}, \bibinfo{person}{Ion Stoica}, {and} \bibinfo{person}{et al.}} \bibinfo{year}{2024}\natexlab{}.
\newblock \showarticletitle{Judging LLM-as-a-judge with MT-bench and Chatbot Arena}. In \bibinfo{booktitle}{\emph{Proceedings of the 37th International Conference on Neural Information Processing Systems}} (New Orleans, LA, USA) \emph{(\bibinfo{series}{NIPS '23})}. \bibinfo{publisher}{Curran Associates Inc.}, \bibinfo{address}{Red Hook, NY, USA}, Article \bibinfo{articleno}{2020}, \bibinfo{numpages}{29}~pages.
\newblock


\bibitem[Zhu et~al\mbox{.}(2023)]%
        {LogHub}
\bibfield{author}{\bibinfo{person}{Jieming Zhu}, \bibinfo{person}{Shilin He}, \bibinfo{person}{Pinjia He}, \bibinfo{person}{Jinyang Liu}, {and} \bibinfo{person}{Michael~R. Lyu}.} \bibinfo{year}{2023}\natexlab{}.
\newblock \showarticletitle{Loghub: A Large Collection of System Log Datasets for AI-driven Log Analytics}. In \bibinfo{booktitle}{\emph{2023 IEEE 34th International Symposium on Software Reliability Engineering (ISSRE)}}. \bibinfo{pages}{355--366}.
\newblock
\urldef\tempurl%
\url{https://doi.org/10.1109/ISSRE59848.2023.00071}
\showDOI{\tempurl}


\bibitem[Zhu et~al\mbox{.}(2021)]%
        {unilog}
\bibfield{author}{\bibinfo{person}{Yichen Zhu}, \bibinfo{person}{Weibin Meng}, \bibinfo{person}{Ying Liu}, \bibinfo{person}{Shenglin Zhang}, \bibinfo{person}{Tao Han}, \bibinfo{person}{Shimin Tao}, {and} \bibinfo{person}{Dan Pei}.} \bibinfo{year}{2021}\natexlab{}.
\newblock \bibinfo{title}{UniLog: Deploy One Model and Specialize it for All Log Analysis Tasks}.
\newblock
\newblock
\showeprint[arxiv]{2112.03159}~[cs.NI]
\urldef\tempurl%
\url{https://arxiv.org/abs/2112.03159}
\showURL{%
\tempurl}


\end{thebibliography}

\clearpage
\appendix
\bf{Appendix A: LLM Prompts}
%\vspace{.25in}

\begin{figure}[b]
  \centerline{\includegraphics[width=\linewidth]{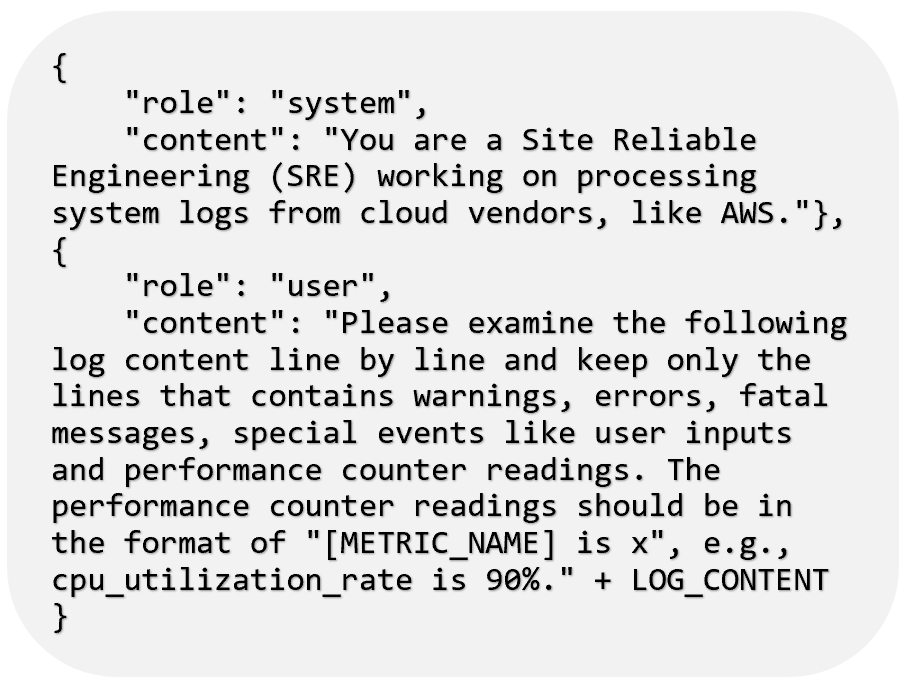}}
  \caption{Prompt for LLM to process log file.}
  \label{fig:log_prompt}
\end{figure}

\begin{figure}[b]
  \centerline{\includegraphics[width=\linewidth]{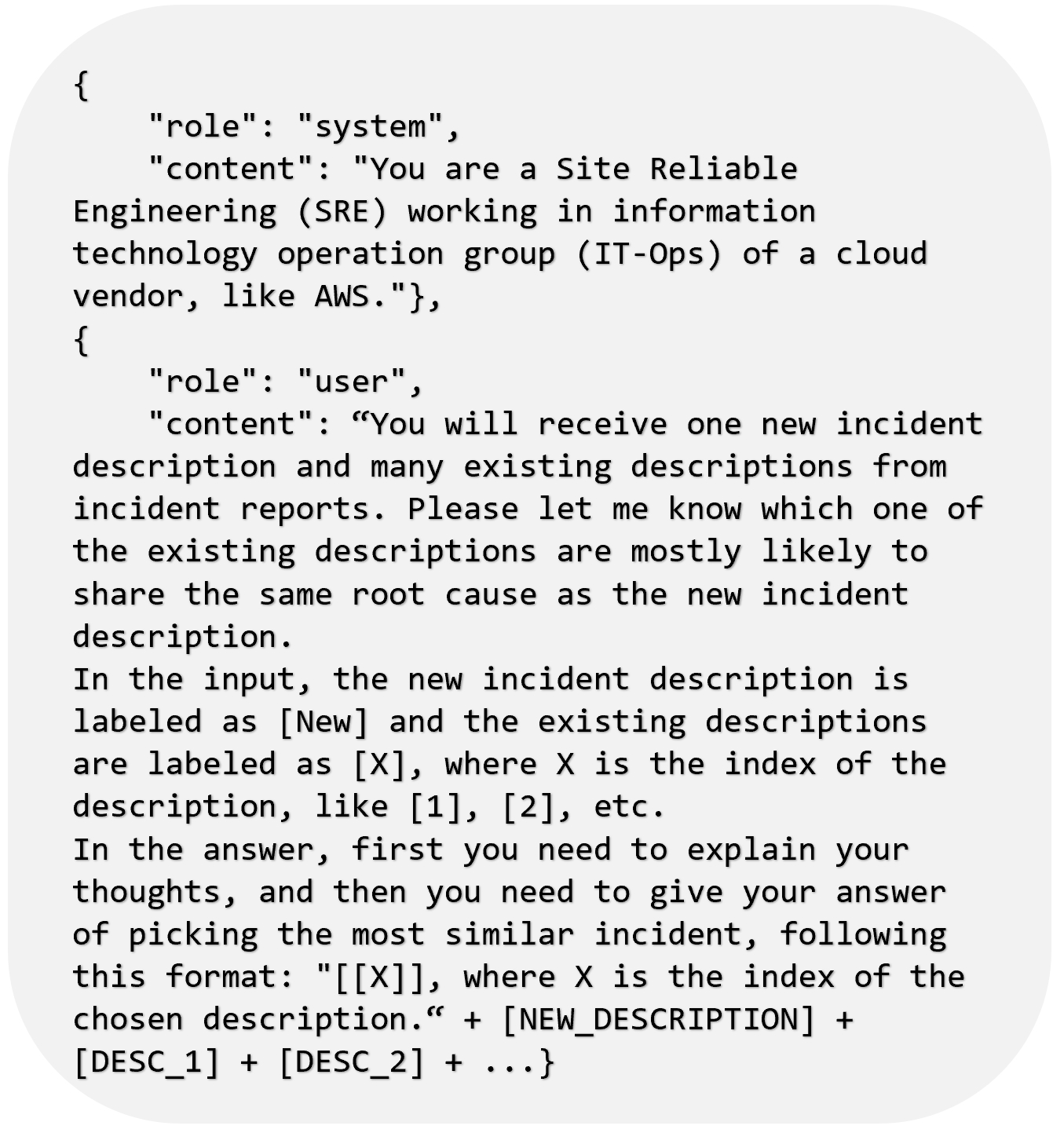}}
  \caption{Prompt for the Evaluation LLM.}
  \label{fig:evaluation_llm}
\end{figure}

\begin{figure}[b]
  \centerline{\includegraphics[width=\linewidth]{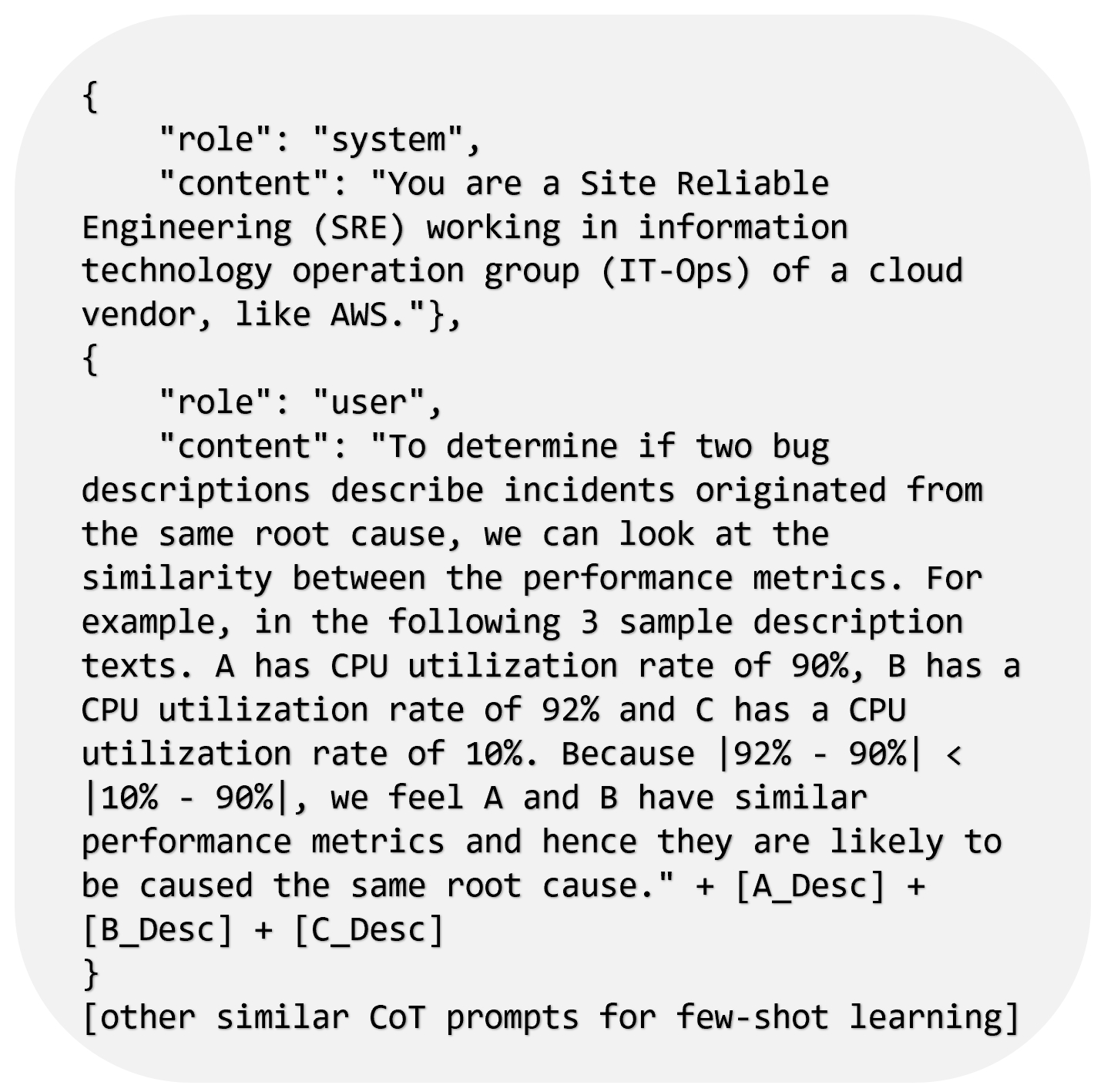}}
  \caption{The CoT contexts for the Evaluation LLM.}
  \label{fig:evaluation_cot}
\end{figure}

\begin{figure}[b]
  \centerline{\includegraphics[width=\linewidth]{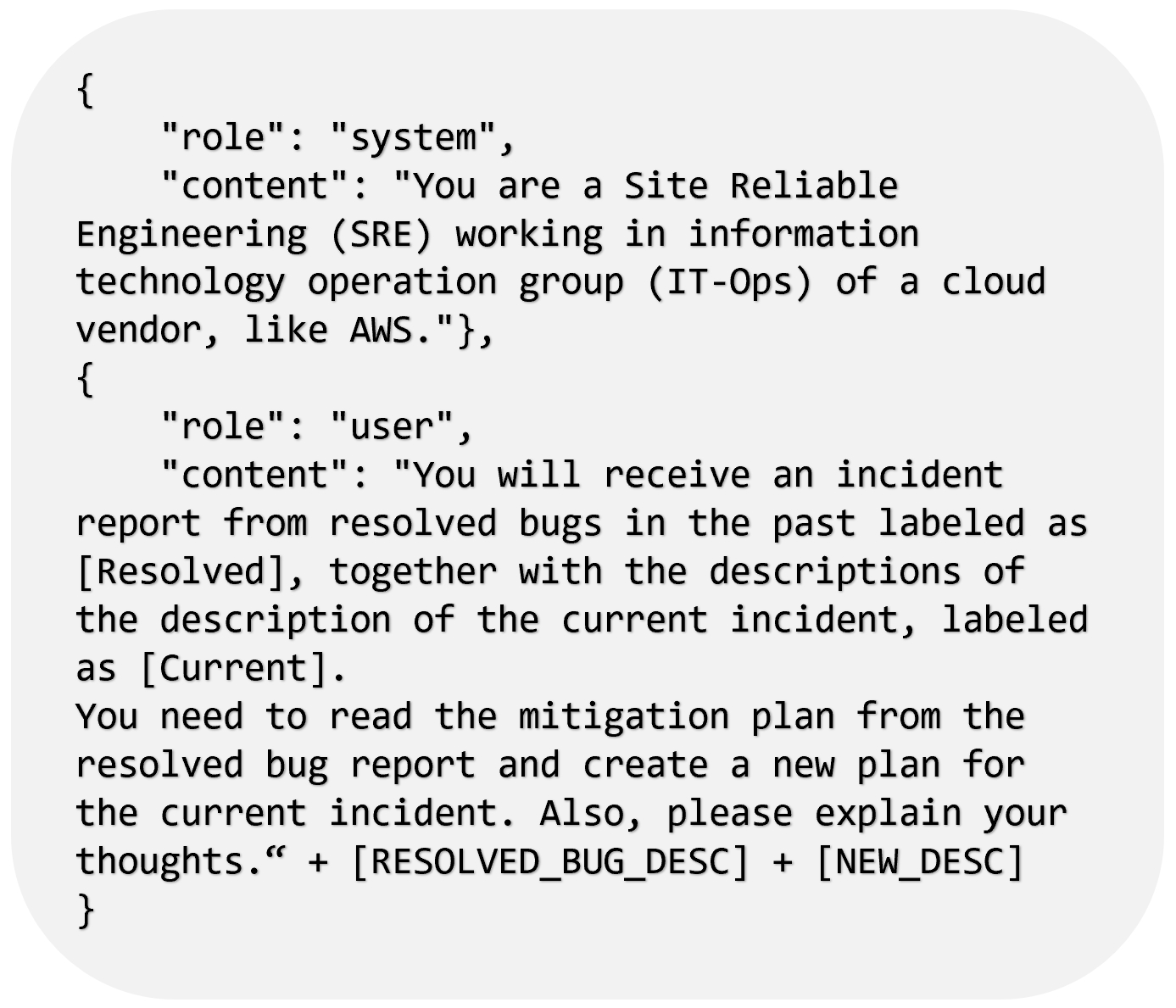}}
  \caption{Prompt for the Generator LLM.}
  \label{fig:generate}
\end{figure}
\end{document}